\documentclass[a4paper, 10pt]{article}

\usepackage{epsfig}
\usepackage{amsmath}
\usepackage{amssymb}
\usepackage[english,ruled,vlined]{algorithm2e}

\newcommand{\bel}{\mathrm{bel}}
\newcommand{\pl}{\mathrm{pl}}
\newcommand{\betP}{\mathrm{betP}}
\newcommand{\Bel}{\mathrm{Bel}}
\newcommand{\Pl}{\mathrm{Pl}}
\newcommand{\GPT}{\mathrm{GPT}}
\newcommand{\eR}{\mbox{I\hspace{-.15em}R}}
\newcommand{\ind}{\mbox{1\hspace{-.25em}l}}
\newcommand{\capop}{\operatornamewithlimits{\cap}}
\newcommand{\PCRmo}{\mathrm{PCR6}}
\newcommand{\PCRdsm}{\mathrm{PCR5}}
\newcommand{\Pmof}{\mathrm{PCR6f}}
\newcommand{\Pmog}{\mathrm{PCR6g}}
\newcommand{\conj}{\mathrm{c}}
\newcommand{\DP}{\mathrm{DP}}

\title{A new generalization of the proportional conflict redistribution rule stable in terms of decision}
\author{Arnaud Martin and Christophe Osswald}

\begin{document}
\maketitle

\begin{abstract}
In this chapter, we present and discuss a new generalized proportional conflict redistribution rule. The Dezert-Smarandache extension of the Demster-Shafer theory has relaunched the studies on the combination rules especially for the management of the conflict. Many combination rules have been proposed in the last few years. We study here different combination rules and compare them in terms of decision on didactic example and on generated data. Indeed, in real applications, we need a reliable decision and it is the final results that matter. This chapter shows that a fine proportional conflict redistribution rule must be preferred for the combination in the belief function theory.

{\bf Keywords:} Experts fusion, DST, DSmT, generalized PCR, DSmH.
\end{abstract}

\section{Introduction}

Many fusion theories have been studied for the combination of the experts opinions such as voting rules \cite{Xu92,Lam97}, possibility theory \cite{Zadeh78,Dubois88a}, and belief function theory \cite{Dempster67,Shafer76}. We can divide all these fusion approaches into four steps: the modelization, the parameters estimation depending on the model (not always necessary), the combination, and the decision. The most difficult step is presumably the first one. If both possibility and probability-based theories can modelize imprecise and uncertain data at the same time, in a lot of applications, experts can express their certitude on their perception of the reality. As a result, probabilities theory such as the belief function theory is more adapted. In the context of the belief function theory, the Dempster-Shafer theory (DST) \cite{Dempster67, Shafer76} is based on the use of functions defined on the power set $2^\Theta$ (that is the set of all the disjunctions of the elements of $\Theta$). Hence the experts can express their opinion not only on $\Theta$ but also on $2^\Theta$ as in the probabilities theory. The extension of this power set into the hyper power set $D^\Theta$ (that is the set of all the disjunctions and conjunctions of the elements of $\Theta$) proposed by Dezert and Smarandache \cite{Dezert02}, gives more freedom to the expert. This extension of the DST is called Dezert-Smarandache Theory (DSmT).

This extension has relaunched the studies on the combination rules. The combination of multiple sources of information has still been an important subject of research since the proposed combination rule given by Dempster \cite{Dempster67}. Hence, many solutions have been studied in order to manage the conflict \cite{Yager87,Dubois88,Inagaki91,Smets93,Smets97,Lefevre02a,Lefevre02b,Smarandache05, Florea06}. These combination rules are the most of time compared following the properties of the operator such as associativity, commutativity, linearity, anonymity and on special and simple cases of experts responses \cite{Smets93, Sun04, Daniel04}. 

In real applications, we need a reliable decision and it is the final results that matter. Hence, for a given application, the best combination rule is the rule given the best results. For the decision step, different functions such as credibility, plausibility and pignistic probability \cite{Shafer76, Smets90, Dezert04} are usually used. 

In this chapter, we discuss and compare different combination rules especially managing the conflict. First, the principles of the DST and DSmT are recalled. We present the formalization of the belief function models, different rules of combination and decision. One the combination rule (PCR5) proposed by \cite{Smarandache05} for two experts is mathematically one of the best for the proportional redistribution of the conflict applicable in the context of the DST and the DSmT. In the section \ref{PCR6}, we propose a new extension of this rule for more experts, the PCR6 rule. This new rule is compared to the generalized PCR5 rule given in \cite{Dezert06}, in the section \ref{discussion}. Then this section presents a comparison of different combination rules in terms of decision in a general case, where the experts opinions are randomly simulated. We demonstrate also that some combination rules are different in terms of decision, in the case of two experts and two classes.

\section{Theory Bases}
\subsection{Belief Function Models}
The belief functions or basic belief assignments $m$ are defined by the mapping of the power set $2^\Theta$ onto $[0,1]$, in the DST, and by the mapping of the hyper-power set $D^\Theta$ onto $[0,1]$, in the DSmT, with:

\begin{equation}
\label{close}
m(\emptyset)=0,
\end{equation}
and 
\begin{equation}
\label{normDST}
\sum_{X\in 2^\Theta} m(X)=1,
\end{equation}
in the DST, and 
\begin{equation}
\label{normDSmT}
\sum_{X\in D^\Theta} m(X)=1,
\end{equation}
in the DSmT. 

The equation (\ref{close}) is the hypothesis at a closed world \cite{Shafer76, Smarandache04}. We can define the belief function only with:
\begin{equation}
\label{open}
m(\emptyset)>0,
\end{equation}
and the world is open \cite{Smets90}. In a closed world, we can also add one element in order to propose an open world.

These simple conditions in equation (\ref{close}) and (\ref{normDST}) or (\ref{close}) and (\ref{normDSmT}), give a large panel of definitions of the belief functions, which is one the difficulties of the theory. The belief functions must therefore be chosen according to the intended application. 

\subsection{Combination rules}
Many combination rules have been proposed in the last few years in the context of the belief function theory (\cite{Yager87, Dubois88, Smets90, Smets93, Smarandache04, Smarandache05}, {\it etc.}). In the context of the DST, the combination rule most used today seems to be the conjunctive rule given by \cite{Smets90} for all $X \in 2^\Theta$ by:
\begin{eqnarray}
\label{conjunctive}
m_\conj(X)=\sum_{Y_1 \cap ... \cap Y_M = X} \prod_{j=1}^M m_j(Y_j),
\end{eqnarray}
where $Y_j \in 2^\Theta$ is the response of the expert $j$, and $m_j(Y_j)$ the associated belief function.

However, the conflict can be redistributed on partial ignorance like in the Dubois and Prade rule \cite{Dubois88}, a mixed conjunctive and disjunctive rule given for all $X \in 2^\Theta$, $X\neq \emptyset$ by: 
\begin{eqnarray}
\label{DP}
m_\DP(X)=\sum_{Y_1 \cap ... \cap Y_M = X} \prod_{j=1}^M m_j(Y_j)+\sum_{
\begin{array}{c}
Y_1 \cup ... \cup Y_M = X\\
Y_1 \cap ... \cap Y_M = \emptyset \\
\end{array}} \prod_{j=1}^M m_j(Y_j),
\end{eqnarray}
where $Y_j \in 2^\Theta$ is the response of the expert $j$, and $m_j(Y_j)$ the associated belief function.

In the context of the DSmT, the conjunctive rule can be used for all $X \in D^\Theta$ and $Y \in D^\Theta$. The rule given by the equation (\ref{DP}), called DSmH \cite{Smarandache04}, can be write in $D^\Theta$ for all $X \in D^\Theta$, $X\not\equiv \emptyset$ \footnote{The notation $X\not\equiv \emptyset$ means that $X \neq \emptyset$ and following the chosen model in $D^\Theta$, $X$ is not one of the element of $D^\Theta$ defined as $\emptyset$. For example, if $\Theta=\{A, B, C\}$, we can define a model for which the expert can provide a mass on $A\cap B$ and not on $A \cap C$, so $A\cap B\neq \emptyset$ and $A\cap B=\emptyset$} by:
\begin{equation}
  \label{DSmH}
  \begin{array}{c}
    m_H(X)=\displaystyle \sum_{Y_1 \cap ... \cap Y_M = X} \prod_{j=1}^M m_j(Y_j)+\displaystyle \sum_{
      \begin{array}{c}
	\scriptstyle Y_1 \cup ... \cup Y_M = X\\
	\scriptstyle Y_1 \cap ... \cap Y_M \equiv \emptyset \\
    \end{array}} 
    \prod_{j=1}^M m_j(Y_j)+\\ 
    \displaystyle \sum_{
      \begin{array}{c}
	\scriptstyle \left\{u(Y_1) \cup ... \cup u(Y_M) = X\right\} \\ 
	\scriptstyle Y_1, ..., Y_M \equiv \emptyset \\
    \end{array}} \displaystyle \!\!\!\!\prod_{j=1}^M m_j(Y_j)+
    \!\!\!\!\!\!\!\!\displaystyle \sum_{
      \begin{array}{c}
	\scriptstyle \left\{ u(Y_1) \cup ... \cup u(Y_M) \equiv \emptyset \, \mbox{and} \, X=\Theta\right\} \\
	\scriptstyle Y_1, ..., Y_M \equiv \emptyset \\
    \end{array}} \displaystyle \prod_{j=1}^M m_j(Y_j),
  \end{array}
\end{equation}
where $Y_j \in D^\Theta$ is the response of the expert $j$, $m_j(Y_j)$ the associated belief function, and $u(Y)$ is the function giving the union that compose $Y$ \cite{Smarandache04chap4}. For example if $Y=(A\cap B) \cup (A \cap C)$, $u(Y)=A \cup B \cup C$.

If we want to take the decision only on the elements in $\Theta$, some rules propose to redistribute the conflict on these elements. The most accomplished is the PCR5 given in \cite{Smarandache05} for two experts and for $X\in D^\Theta$, $X\neq \emptyset$ by:
\begin{eqnarray}
\label{DSmTcombination}
\begin{array}{l}
m_{PCR5}(X)=m_c(X)+\\
\displaystyle
\sum_{\begin{array}{l}
\scriptstyle Y\in D^\Theta, \\
\scriptstyle X\cap Y \equiv \emptyset 
\end{array}} \left(\frac{m_1(X)^2 m_2(Y)}{m_1(X)+m_2(Y)}+\frac{m_2(X)^2 m_1(Y)}{m_2(X)+m_1(Y)}\right),
\end{array}
\end{eqnarray}
where $m_c(.)$ is the conjunctive rule given by the equation (\ref{conjunctive}). 

Note that more rules managing the conflict have been proposed \cite{Yager87,Inagaki91,Lefevre02a,Lefevre02b,Smarandache05, Florea06}.

The comparison of all the combination rules is not the scope of this paper. 

\subsection{Decision rules}

The decision is a difficult task. No measures are able to provide the best decision in all the cases. Generally, we consider the maximum of one of the three functions: credibility, plausibility, and pignistic probability. 

In the context of the DST, the credibility function is given for all $X \in 2^\Theta$ by:
\begin{eqnarray}
\bel(X)=\sum_{Y \in 2^X, Y \neq \emptyset} m(Y).
\end{eqnarray}
The plausibility function is given for all $X \in 2^\Theta$ by:
\begin{eqnarray}
\pl(X)=\sum_{Y \in 2^\Theta, Y\cap X \neq \emptyset} m(Y)=\bel(\Theta)-\bel(X^c),
\end{eqnarray}
where $X^c$ is the complementary of $X$. The pignistic probability, introduced by \cite{Smets90b}, is here given for all $X \in 2^\Theta$, with $X \neq \emptyset$ by:
\begin{eqnarray}
\betP(X)=\sum_{Y \in 2^\Theta, Y \neq \emptyset} \frac{|X \cap Y|}{|Y|} \frac{m(Y)}{1-m(\emptyset)}.
\end{eqnarray}
Generally the maximum of these functions is taken on the elements in $\Theta$, but we will give the values on all the focal elements.

In the context of the DSmT the corresponding generalized functions have been proposed \cite{Dezert04, Smarandache04}.
The generalized credibility $\Bel$ is defined by:
\begin{eqnarray}
\Bel(X)=\sum_{Y \in D^\Theta, Y\subseteq X, Y \not\equiv \emptyset} m(Y)
\end{eqnarray}
The generalized plausibility $\Pl$ is defined by:
\begin{eqnarray}
\Pl(X)=\sum_{Y \in D^\Theta, X \cap Y\not\equiv \emptyset} m(Y)
\end{eqnarray}
The generalized pignistic probability is given for all $X \in D^\Theta$, with $X \neq \emptyset$ is defined by:
\begin{eqnarray}
\GPT(X)=\sum_{Y \in D^\Theta, Y \not\equiv \emptyset} \frac{{\cal C_M}(X \cap Y)}{{\cal C_M}(Y)} m(Y),
\end{eqnarray}
where ${\cal C_M}(X)$ is the DSm cardinality corresponding to the number of parts of $X$ in the Venn diagram of the problem \cite{Dezert04, Smarandache04}.

If the credibility function provides a pessimist decision, the plausibility function is often too optimist. The pignistic probability is often taken as a compromise. We present the three functions for our models.

\section{The generalized PCR rules}
\label{PCR6}
In the equation (\ref{DSmTcombination}), the PCR5 is given for two experts only. Two extensions for three experts and two classes are given in \cite{Smarandache04b}, and the equation for $M$ experts, for $X\in D^\Theta$, $X\not\equiv \emptyset$ is given in \cite{Dezert06} by:

\begin{eqnarray}
  \label{GenePCR_DS}
  & & \!\!\!\!\!\!\!\!\!\!\!\!\!\!\! \nonumber \displaystyle \!\!\!m_\PCRdsm(X) = m_\conj(X) + \\
  & & \!\!\!\!\!\!\!\!\!\!\!\!\!\!\! \sum_{i=1}^M
  m_i(X) \!\!\!\!\!\!\!\!\!\!\!\!\!\!\!\!\!\!\!\!\! \sum_{\begin{array}{c}
      \scriptstyle (Y_{\sigma_i(1)},...,Y_{\sigma_i(M\!-\!1)})\in
      (D^\Theta)^{M\!-\!1} \\
      \scriptstyle {\displaystyle \mathop{\cap}_{\scriptscriptstyle
	  k=1}^{\scriptscriptstyle M\!-\!1}} Y_{\sigma_i(k)} \cap X
      \equiv \emptyset
  \end{array}}
  \!\!\!\!\!\!\!\!\!\!\!\!\!\!\!\frac{
    \displaystyle \Bigg(\prod_{j=1}^{M\!-\!1} m_{\sigma_i(j)}(Y_{\sigma_i(j)})\ind_{j>i}\Bigg)
    \!\!\!\prod_{Y_{\sigma_i(j)}=X} \!\!\!\!\! m_{\sigma_i(j)}(Y_{\sigma_i(j)})
  }{
    \displaystyle \!\!\!\!\!\!\!\!\!\!\!\!\sum_{~~~~~\renewcommand{\arraystretch}{1.8}\begin{array}{c}
	\scriptstyle Z\in\{X, Y_{\sigma_i(1)}, \ldots, Y_{\sigma_i(M\!-\!1)}\}
    \end{array}}
    \!\!\!\!\!\!\!\!\!\!\!\!\!\!\!\!\!\!\!\!\!\!\!\!\!\!\!\!\!\!\!\!\!\prod_{Y_{\sigma_i(j)}=Z}^{\:}\!\!\!\!\!\!\!
    \big(m_{\sigma_i(j)}(Y_{\sigma_i(j)}).T({\scriptstyle
    X\!=\!Z,m_i(X)})\big)},
\end{eqnarray}
where $\sigma_i$ counts from 1 to $M$ avoiding $i$:
\begin{eqnarray}
\label{sigma}
\left\{
\begin{array}{ll}
\sigma_i(j)=j &\mbox{if~} j<i,\\
\sigma_i(j)=j+1 &\mbox{if~} j\geq i,\\
\end{array}
\right.
\end{eqnarray}
and:
\begin{eqnarray}
\label{T}
\left\{
\begin{array}{ll}
T(B,x)=x &\mbox{if $B$ is true},\\
T(B,x)=1 &\mbox{if $B$ is false}.\\
\end{array}
\right.
\end{eqnarray}

We propose another generalization of the equation (\ref{DSmTcombination}) for $M$ experts, for $X\in D^\Theta$, $X\neq \emptyset$:

\begin{eqnarray}
\label{GeneDSmTcombination}
  \displaystyle m_\PCRmo(X) & = & \displaystyle m_\conj(X) + \\
  \nonumber & & \!\!\!\!\!\!\!\!\!\!\sum_{i=1}^M m_i(X)^2
  \!\!\!\!\!\!\!\!\!\!\!\!\!\!\!\!\!\!\!\! \sum_{\begin{array}{c}
      \scriptstyle {\displaystyle \mathop{\cap}_{k=1}^{M\!-\!1}} Y_{\sigma_i(k)} \cap X \equiv \emptyset \\
      \scriptstyle (Y_{\sigma_i(1)},...,Y_{\sigma_i(M\!-\!1)})\in (D^\Theta)^{M\!-\!1}
  \end{array}}
  \!\!\!\!\!
  \left(\!\!\frac{\displaystyle \prod_{j=1}^{M\!-\!1} m_{\sigma_i(j)}(Y_{\sigma_i(j)})}
       {\displaystyle m_i(X) \!+\! \sum_{j=1}^{M\!-\!1}
  m_{\sigma_i(j)}(Y_{\sigma_i(j)})}\!\!\right)\!\!,
\end{eqnarray}
where $\sigma$ is defined like in (\ref{sigma}).

$m_i(X)+\displaystyle \sum_{j=1}^{M-1} m_{\sigma_i(j)}(Y_{\sigma_i(j)}) \neq 0$, $m_c$ is the conjunctive consensus rule given by the equation (\ref{conjunctive}). 

We can propose two more generalized rules given by:
\begin{eqnarray}
  \label{GenePCR_f}
  \displaystyle m_{\Pmof}(X) & = & \displaystyle m_\conj(X) + \\
  \nonumber & &
  \!\!\!\!\!\!\!\!\!\!\!\!\!\!\!\!\!\!\!\!\!\!\!\!\!\!\!\!\!\!\!\!\!\!\!\!\!\!\!\!\!\!\!\!\!\!\!\!
  \sum_{i=1}^M  m_i(X)f(m_i(X)) 
  \!\!\!\!\!\!\!\!\!\!\!\!\!\!\!\!\!\!\!\!\!\!\!
  \displaystyle \sum_{\begin{array}{c}
	\scriptstyle {\displaystyle \mathop{\cap}_{k=1}^{M\!-\!1}} Y_{\sigma_i(k)} \cap X \equiv \emptyset \\
	\scriptstyle (Y_{\sigma_i(1)},...,Y_{\sigma_i(M\!-\!1)})\in (D^\Theta)^{M\!-\!1}
    \end{array}}
    \!\!\!\!\!\!
    \left(\!\!\frac{\displaystyle \prod_{j=1}^{M\!-\!1} m_{\sigma_i(j)}(Y_{\sigma_i(j)})}
	 {\displaystyle f(m_i(X)) \!+\! \sum_{j=1}^{M\!-\!1} f(m_{\sigma_i(j)}(Y_{\sigma_i(j)}))}\!\!\right)\!\!,
\end{eqnarray}
with the same notations that in the equation (\ref{GeneDSmTcombination}), and $f$ an increasing function defined by the mapping of $[0,1]$ onto $\eR^+$. 

The second generalized rule is given by:
\begin{equation}
  \label{GenePCR_g}
  \begin{array}{l}
    \displaystyle m_{\Pmog}(X) =  m_\conj(X) + 
    \displaystyle \sum_{i=1}^M \!\!\!\!\!\!\!\!\!\!\!\!
    \sum_{\begin{array}{c}
	\scriptstyle {\displaystyle \mathop{\cap}_{\scriptscriptstyle
	    k=1}^{\scriptscriptstyle M\!-\!1}} Y_{\sigma_i(k)} \cap X
	\equiv \emptyset\\ 
	\scriptstyle (Y_{\sigma_i(1)},...,Y_{\sigma_i(M\!-\!1)})\in (D^\Theta)^{M\!-\!1}
    \end{array}}\\ \\
     m_i(X) \frac{\displaystyle
      \Bigg(\!\prod_{j=1}^{M\!-\!1}\!\!m_{\sigma_i(j)}(Y_{\sigma_i(j)})\!\!\Bigg)
      \!\!\Bigg(\!\!\!\!\!\!\!\prod_{~~~Y_{\sigma_i(j)}=X}\!\!\!\!\!\!\!\!\!\!\!\ind_{j>i}\!\!\Bigg)
      g\Bigg(\!\!m_i(X)\!\!+\!\!\!\!\!\!\!\!\sum_{Y_{\sigma_i(j)}=X}\displaystyle \!\!\!\!\!\!\!\!m_{\sigma_i(j)}(Y_{\sigma_i(j)})\!\!\!\Bigg)}
	{\displaystyle \sum_{
	    \scriptstyle Z\in\{X, Y_{\sigma_i(1)}, \ldots, Y_{\sigma_i(M\!-\!1)}\}}
	  \!\!\!g\!\!\left(\!\sum_{Y_{\sigma_i(j)}=Z}\!\!\!\!\!\displaystyle m_{\sigma_i(j)}(Y_{\sigma_i(j)})+m_i(X)\ind_{\scriptscriptstyle X=Z}\!\!\!\right)},
  \end{array}
\end{equation}
with the same notations that in the equation (\ref{GeneDSmTcombination}), and $g$ an increasing function defined by the mapping of $[0,1]$ onto $\eR^+$.

For instance, we can choose $f(x)=g(x)=x^\alpha$, with $\alpha \in \eR^+$. 

Algorithms for the Dubois and Prade (equation (\ref{DP})), the $\PCRdsm$ (equation (\ref{GenePCR_DS})), the $\PCRmo$ (equation (\ref{GeneDSmTcombination})), the $\Pmof$ (equation (\ref{GenePCR_f})), and the $\Pmog$ (equation (\ref{GenePCR_g})) combinations are given in appendix.

\paragraph{Remarks on the generalized PCR rules}
\begin{itemize}

\item $\displaystyle \capop_{k=1}^{M-1} Y_k \cap X \equiv \emptyset$ means that
  $\displaystyle \capop_{k=1}^{M-1} Y_k \cap X$ is considered as a conflict by the
  model: $\displaystyle m_i(X)\prod_{k=1}^{M-1}m_{\sigma_i(k)}(Y_{\sigma_i(k)})$ has to
  be redistributed on $X$ and the $Y_k$.
\item The second term of the equation (\ref{GeneDSmTcombination}) is
null if $\displaystyle \capop_{k=1}^{M-1} Y_k \cap X \not\equiv \emptyset$, hence in a
general model in $D^\Theta$ for all $X$ and $Y \in D^\Theta$, $X\cap Y
\neq \emptyset$. The PCR5 and PCR6 are exactly the conjunctive rule:
there is never any conflict. However in $2^{2^{\Theta}}$, $\exists X,
Y \in 2^{2^{\Theta}}$ such as $X\cap Y=\emptyset$.

\item One of the principal problem of the PCR5 and PCR6 rules is the non
associativity. That is is a real problem for dynamic fusion. Take for
example three experts and two classes giving:

\begin{center}
  \begin{tabular}{|l|c|c|c|c|}
    \hline
    & $\emptyset$ & $A$ & $B$ & $\Theta$ \\
    \hline
    Expert 1 & 0 & 1 & 0 & 0 \\
    \hline
    Expert 2 & 0 & 0 & 1 & 0\\
    \hline
    Expert 3 & 0 & 0 & 1 & 0 \\
    \hline
  \end{tabular}
\end{center}

If we fuse the expert 1 and 2 and then 3, the PCR5 and the PCR6 rules give:
$$m_{12}(A) = 0.5, ~~ m_{12}(B) = 0.5,$$
and
$$m_{(12)3}(A) = 0.25, ~~ m_{(12)3}(B) = 0.75.$$

Now if we fuse the expert 2 and 3 and then 1, the PCR5 and the PCR6 rules give:
$$m_{23}(A) = 0, ~~ m_{23}(B) = 1,$$
and
$$m_{(12)3}(A) = 0.5, ~~ m_{(12)3}(B) = 0.5,$$
and the result is not the same.

With the generalized PCR6 rule we obtain:
$$m_{(123)}(A) = 1/3, ~~ m_{(123)}(B) = 2/3,$$
a more intuitive result.

\item The conflict is not only redistributed on singletons. For example if three experts give:
\begin{center}
  \begin{tabular}{|l|c|c|c|c|}
    \hline
    & $A\cup B$ & $B\cup C$ & $A\cup C$ & $\Theta$ \\
    \hline
    Expert 1  & 0.7 & 0 & 0 & 0.3 \\
    \hline
    Expert 2 & 0 & 0 & 0.6 & 0.4 \\
    \hline
    Expert 3 & 0 & 0.5 & 0 & 0.5 \\
    \hline
  \end{tabular}
\end{center}

The conflict is given here by 0.7$\times$0.6$\times$0.5=0.21, with the generalized PCR6 rule we obtain:
$$m_{(123)}(A) = 0.21,$$
$$m_{(123)}(B) = 0.14,$$
$$m_{(123)}(C) = 0.09,$$
$$m_{(123)}(A\cup B) = 0.14+0.21.\frac{7}{18}\simeq 0.2217,$$
$$m_{(123)}(B \cup C) = 0.06+0.21.\frac{5}{18} \simeq 0.1183,$$
$$m_{(123)}(A \cup C) = 0.09+0.21.\frac{6}{18} = 0.16,$$
$$m_{(123)}(\Theta) = 0.06.$$

\end{itemize}

\section{Discussion on the decision following the combination rules}
\label{discussion}

In order to compare the previous rules in this section, we study the decision on the basic belief assignments obtained by the combination. Hence, we consider here the induced order on the singleton given by the plausibility, credibility, pignistic probability functions, or directly by the masses. Indeed, in order to compare the combination rules, we think that the study on the induced order of these functions is more informative than the obtained masses values. All the combination rules presented here are not idempotent, for instance for the conjunctive non-normalized rule: 
\begin{center}
  \begin{tabular}{|c|c|c|c|c|}
		\hline
		& $\emptyset$ & $A$ & $B$ & $C$ \\
		\hline
		$m_1$ & 0& 0.6 & 0.3 & 0.1 \\
		\hline
		$m_1$ & 0& 0.6 & 0.3 & 0.1 \\
		\hline
		$m_{11}$ & 0.54 & 0.36 & 0.09 & 0.01 \\
		\hline
\end{tabular}
\end{center}

So, if we only compare the rules looking the obtained masses, we have normalize them with the auto-conflict given by the combination of a mass with itself. However, if $m_1(A) > m_1(B)$, then $m_{11}(A)>m_{11}(B)$.

\subsection{Extending the PCR rule for more than two experts}

In \cite{Smarandache04b}, two approaches are presented in order to extend the PCR5 rule. The second approach suggests to fuse the first two experts and then fuse the third expert. However the solution depend on the order of the experts because of the non-associativity of the rule, and so it is not satisfying. 

The first approach proposed in \cite{Smarandache04b}, that is the equation (\ref{GenePCR_DS}) proposes to redistribute the conflict about the singleton, {\em e.g.} if we have $m_1(A)m_3(B)m_2(A\cup B)$, the conflict is redistributed on $A$ and $B$ proportionally to $m_1(A)$ and $m_3(B)$. But this approach do not give solution if we have for instance $m_1(A\cup B)m_2(B\cup C)m_3(A\cup C)$ where the conflict is $A\cap B \cap C$ and we have no idea on the masses for $A$, $B$ and $C$. 

Moreover, if we have $m_(A)m_2(B)m_3(B)$ the proposed solution distribute the conflict to $A$ and $B$ with respect to $m_1(A)$ and $m_2(B)m_3(B)$ and not $m_2(B)+m_3(B)$ that is more intuitive. For example, if $m_1(A)=m_2(B)=m_3(B)=0.5$, 0.0833 and 0.0416 is added to the masses $A$ and $B$ respectively, while there is more consensus on $B$ than on $A$ and we would expected the contrary: 0.0416 and 0.0833 could be added to the masses $A$ and $B$ respectively.

What is more surprising are the results given by PCR5 and PCR6 on the
following example:
\begin{center}
\begin{tabular}{|l|c|c|c|c|c|c|c|}
\hline
& A & B & C & D & E & F & G \\
\hline
Expert 1 &  0.0 & 0.57 & 0.43 & 0.0 & 0.0 & 0.0 & 0.0 \\
Expert 2 &  0.58 & 0.0 & 0.0 & 0.42 & 0.0 & 0.0 & 0.0 \\
Expert 3 &  0.58 & 0.0 & 0.0 & 0.0 & 0.42 & 0.0 & 0.0 \\
Expert 4 &  0.58 & 0.0 & 0.0 & 0.0 & 0.0 & 0.42 & 0.0 \\
Expert 5 &  0.58 & 0.0 & 0.0 & 0.0 & 0.0 & 0.0 & 0.42 \\
\hline
\end{tabular}
\end{center}

As all the masses are on singletons, neither PCR5 nor PCR6 can put any
mass on total or partial ignorance. So the fusion result is always a
probability, and $\bel(X)=\betP(X)=\pl(X)$.

Conflict is total: conjunctive rule does not provide any
information. PCR5 and PCR6 give the following results:

\begin{center}
\begin{tabular}{|l|c|c|c|c|c|c|c|}
\hline
& A & B & C & D & E & F & G \\
\hline
PCR5 & 0.1915 & 0.2376 & 0.1542 & 0.1042 & 0.1042 & 0.1042 & 0.1042 \\
\hline
PCR6 & 0.5138 & 0.1244 & 0.0748 & 0.0718 & 0.0718 & 0.0718 & 0.0718 \\
\hline
\end{tabular}
\end{center}

So decision is ``A'' according to PCR6, and decision is ``B''
according to PCR5. However, for any subset of 2, 3 or 4 experts,
decision is ``A'' for any of these combination rules.

\subsection{Stability of decision process}

The space where experts can define their opinions on which $n$ classes are present in a given tile is a part of $[0,1]^n$: $\mathcal{E} = [0,1]^n\cap \left\{ (x_1,...,x_n) \in \eR / \displaystyle \sum_{i=1}^n x_i \leq 1 \right\}$. In order to study the different combination rules, and the situations where they differ, we use a Monte Carlo method, considering the masses given on each class ($a_X$) by each expert, as uniform variables, filtering them by the condition $\displaystyle \sum_{X\in\Theta}a_X\leq 1$ for one expert.

Thus, we measure the proportion of situations where decision differs between the conjunctive combination rule, and the PCR, where conflict is proportionally distributed.

We can not choose $A\cap B$, as the measure of $A\cap B$ is always lower (or equal with probability 0) than the measure of $A$ or $B$. In the case of two classes, $A\cup B$ is the ignorance, and is usually excluded (as it always maximizes $\bel$, $\pl$, $\betP$, $\Bel$, $\Pl$ and $\GPT$). We restrict the possible choices to singletons, $A$, $B$, etc. Therefore, it is equivalent to tag the tile by the most credible class (maximal for $\bel$), the most  plausible (maximal for $\pl$), the most probable (maximal for $\betP$) or the heaviest (maximal for $m$), as the only focal elements are singletons, $\Theta$ and $\emptyset$.

The only situation where the total order induced by the masses $m$ on singletons can be modified is when the conflict is distributed on the
singletons, as is the case in the PCR method. \\

Thus, for different numbers of classes, the decision obtained by fusing the experts' opinions is much less stable:

\begin{center}
  \begin{tabular}{|l|c|c|c|c|c|c|}
    \hline
    number of classes & 2 & 3 & 4 & 5 & 6 & 7 \\
    \hline
    \multicolumn{7}{|l|}{decision change in the two experts case} \\
    \hline
     PCR/DST & 0.61\% & 5.51\% & 9.13\% & 12.11\% & 14.55\% & 16.7\% \\
     PCR/DP & 0.61\% & 2.25\% & 3.42\% & 4.35\% & 5.05\% & 5.7\% \\
     DP/DST & 0.00\% & 3.56\% & 6.19\% & 8.39\% & 10.26\% & 11.9\% \\     
    \hline
    \multicolumn{7}{|l|}{decision change in the three experts case} \\
    \hline
     PCR6/DST & 1.04\% & 8.34\% & 13.90\% & 18.38\% & 21.98\% & 25.1\% \\    
     PCR6/DP & 1.04\% & 5.11\% & 7.54\% & 9.23\% & 10.42\% & 11.3\% \\
     DP/DST & 0.00\% & 4.48\% & 8.88\% & 12.88\% & 16.18\% & 19.0\% \\
    \hline
  \end{tabular}
\end{center} 

Therefore, the specificity of PCR6 appears mostly with more than two
classes, and the different combination rules are nearly equivalent
when decision must be taken within two possible classes.

For two experts and two classes, the mixed rule (DP) and the
conjunctive rule (DST) are equivalent. For three experts, we use the
generalized PCR6 (\ref{GeneDSmTcombination}).

The percentage of decision differences defines a distance between
fusion methods: $d(\mbox{PCR6,DST}) \leq d(\mbox{PCR6,DP})+d(\mbox{DP,DST})$. The two other
triangular inequalities are also true. As we have $d(PCR6,DST) \geq
d(\mbox{PCR6,DP})$ and $d(\mbox{PCR,DST}) \geq d(\mbox{DP,DST})$ for any number of experts
or classes, we can conclude that the mixed rule lies between the PCR6
method and the conjunctive rule.

\begin{figure}[htb]
  \begin{center}
    \begin{tabular}{cc}
      \includegraphics[height=5.5cm]{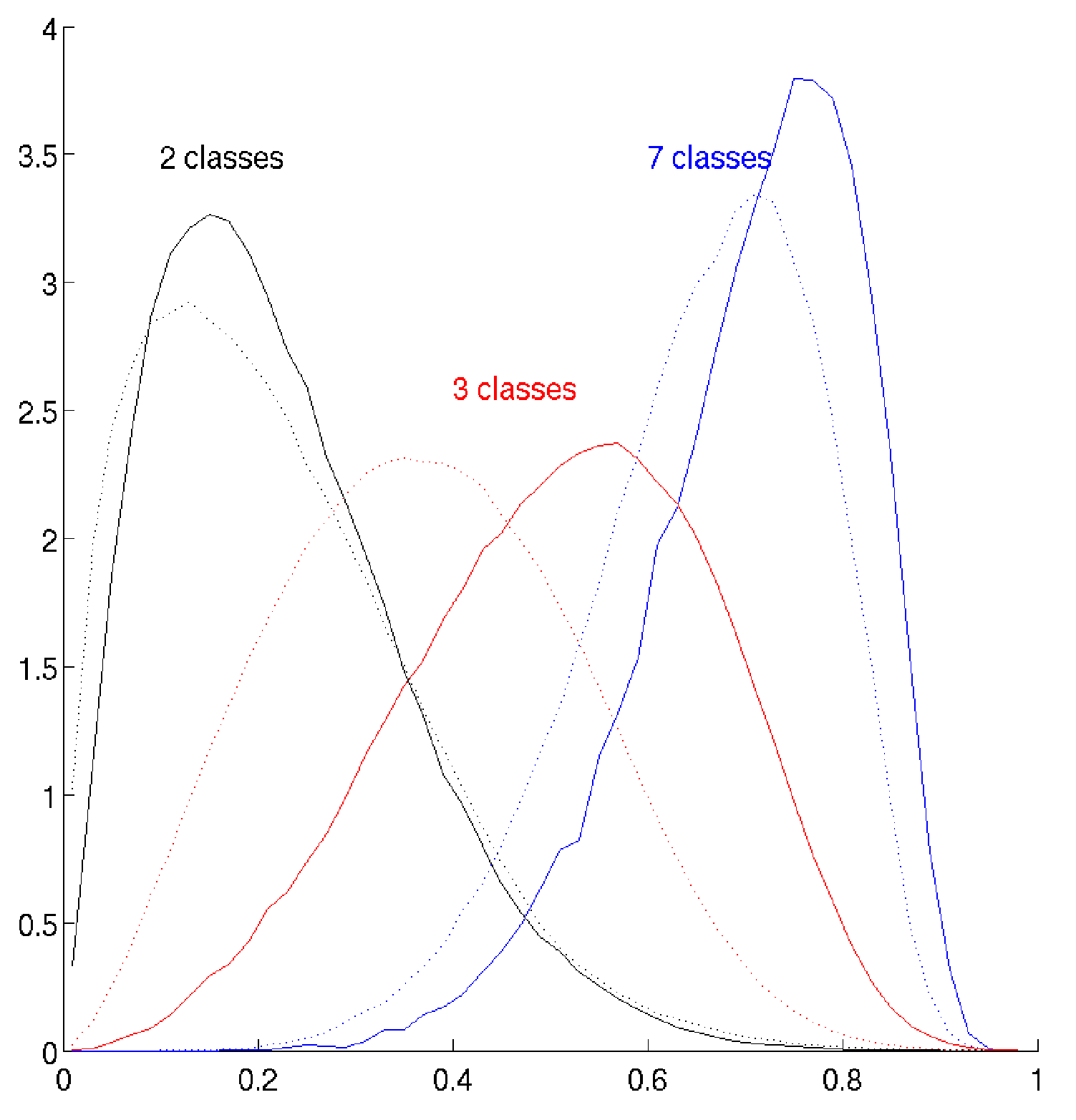} &
      \includegraphics[height=5cm]{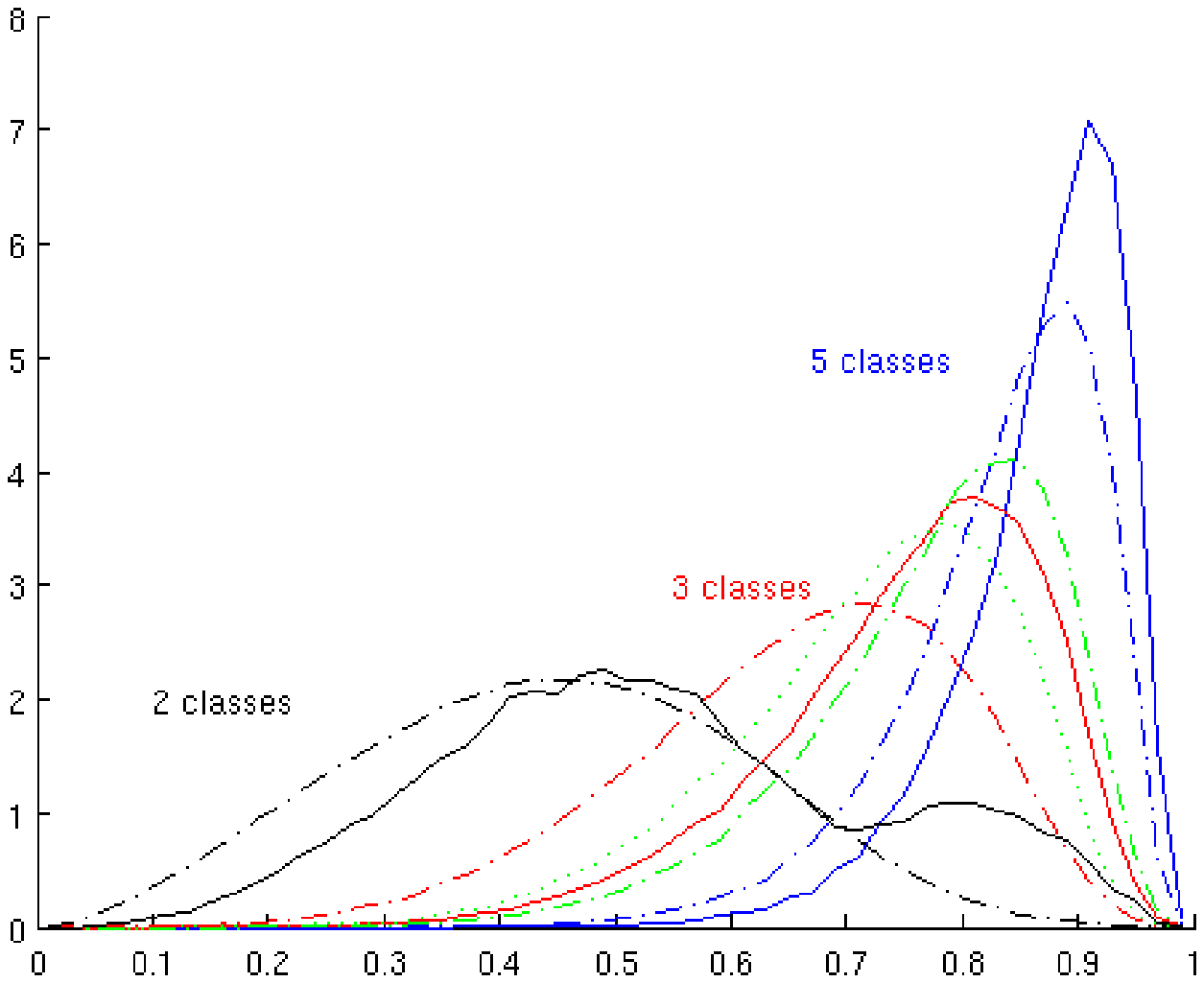} \\
    \end{tabular}
  \end{center}
  \vspace{-0.5cm}
  \caption{Density of conflict for (left) two uniform random experts
    and (right) three uniform random experts; with and without
    decision change}
  \label{conflictdensity}
\end{figure}

The figure \ref{conflictdensity} shows the density of conflict within
$\mathcal{E}$. The left part shows the conflict for two random experts
and a number of classes of 2, 3 or 7. Plain lines show conflict when
there is difference between decisions, and dashed lines show the
overall conflict. Right part shows the conflict values for three
experts; plain lines show the conflict where there is a difference
between the PCR rule and the conjunctive method.

Conflict is more important in this subspace where decision changes
with the method used, mostly because a low conflict usually means a
clear decision. The measure on the best class is often very different
than measure on the second best class.

Dashed green line represents the conflict density for 3 classes when
there is a difference between conjunctive rule and mixed rule. Dotted
green line represents the conflict density for 3 classes when there is
a difference between PCR6 rule and mixed rule. We can see that an high
conflict level emphasizes mostly a decision change between conjunctive
and mixed rule.

\subsection{Calculi for two experts and two classes}

For the ``two experts and two classes'' case, it is difficult to characterize analytically the stability of the decision process between the conjunctive rule and the PCR rule (the PCR5 and PCR6 rules are the same in the two experts case). Note that in this case the DSmH rule given by the equation (\ref{DSmH}), the mixed rule given by the equation (\ref{DP}) and the conjunctive rule given by the equation (\ref{conjunctive}) are equal. However, we can easily resolve few cases where the final decision does not depend on the chosen combination rule.

Standard repartition of expert's opinions is given by this table:
\begin{center}
  \begin{tabular}{|l|c|c|c|c|}
    \hline
    & $\emptyset$ & $A$ & $B$ & $\Theta$ \\
    \hline
    Expert 1 & 0 & $a_1$ & $b_1$ & $1-a_1-b_1$ \\
    \hline
    Expert 2 & 0 & $a_2$ & $b_2$ & $1-a_2-b_2$ \\
    \hline
  \end{tabular}
\end{center}

The conjunctive rule gives:
$$m_c(\emptyset) = a_1b_2+a_2b_1,$$
$$m_c(A) = a_1 + a_2 - a_1a_2 - a_1b_2 - a_2b_1 = a_1 + a_2 - a_1a_2 -
m_c(\emptyset),$$
$$m_c(B) = b_1 + b_2 - b_1b_2 - a_1b_2 - a_2b_1 = b_1 + b_2 - b_1b_2 -
m_c(\emptyset),$$
$$m_c(\Theta) = (1-a_1-b_1)(1-a_2-b_2).$$

PCR gives:
$$m_{PCR}(A) = m(A) + \frac{a_1^2b_2}{a_1+b_2} + \frac{a_2^2b_1}{a_2+b_1},$$
$$m_{PCR}(B) = m(B) + \frac{a_1b_2^2}{a_1+b_2} + \frac{a_2b_1^2}{a_2+b_1},$$
$$m_{PCR}(\emptyset) = 0 ~~~~\mathrm{and}~~~~ m_{PCR}(\Theta) = m_c(\Theta).$$

The stability of the decision is reached if we do not have:
\begin{eqnarray}
\left\{
\begin{array}{l}
m_c(A)>m_c(B)~\mathrm{and}~m_{PCR}(A)<m_{PCR}(B)\\
\mathrm{or}\\
m_c(A)<m_c(B)~\mathrm{and}~m_{PCR}(A)>m_{PCR}(B)\\
\end{array}
\right.
\end{eqnarray}
That means for all $a_1$, $a_2$, $b_1$ and $b_2\in [0,1]$:
\begin{eqnarray}
\begin{array}{l}
\left\{
\begin{array}{l}
a_2+a_1(1-a_2)-b_1(b_2-1)-b_2>0\\
a_1(1-a_2)+a_2\left((1+b_1\left(1-\frac{2}{(1+a_2/b_1)}\right)\right)-b_1(1-b_2)\\
~~~~~~~~~~~~~~~~~~~~-b_2\left(1+a_1\left((1-\frac{2}{(1+b_2/a_1))}\right)\right)<0\\
a_1+b_1 \in [0,1]\\
a_2+b_2 \in [0,1]\\
\end{array}
\right.\\
\mathrm{or}\\
\left\{
\begin{array}{l}
a_2+a_1(1-a_2)-b_1(b_2-1)-b_2<0\\
a_1(1-a_2)+a_2\left((1+b_1\left(1-\frac{2}{(1+a_2/b_1)}\right)\right)-b_1(1-b_2)\\
~~~~~~~~~~~~~~~~~~~~-b_2\left(1+a_1\left((1-\frac{2}{(1+b_2/a_1))}\right)\right)>0\\
a_1+b_1 \in [0,1]\\
a_2+b_2 \in [0,1]\\
\end{array}
\right.
\end{array}
\end{eqnarray}
This system of inequation is difficult to solve, but with the help of a Monte Carlo method, considering the weights $a_1$, $a_2$, $b_1$ and $b_2$, as uniform variables we can estimate the proportion of points ($a_1,a_2,b_1,b_2$) solving this system.

We note that absence of solution in spaces where $a_1+b_1>1$ or $a_2+b_2>1$ comes from the two last conditions of the system. Also there is no solution if $a_1=b_1$ (or $a_2=b_2$ by symmetry) and if $a_1=b_2$ (or $a_2=b_1$ by symmetry). This is proved analytically.

\subsubsection{Case $a_1=b_1$}

In this situation, expert 1 considers that the data unit is equally
filled with classes $A$ and $B$:

\begin{center}
  \begin{tabular}{|l|c|c|c|c|}
    \hline
    & $\emptyset$ & $A$ & $B$ & $\Theta$ \\
    \hline
    Expert 1 & 0 & $x$ & $x$ & $1-2x$ \\
    \hline
    Expert 2 & 0 & $y$ & $z$ & $1-y-z$ \\
    \hline
  \end{tabular}
\end{center}

\begin{figure}[htb]
  \begin{center}
      \includegraphics[height=6cm]{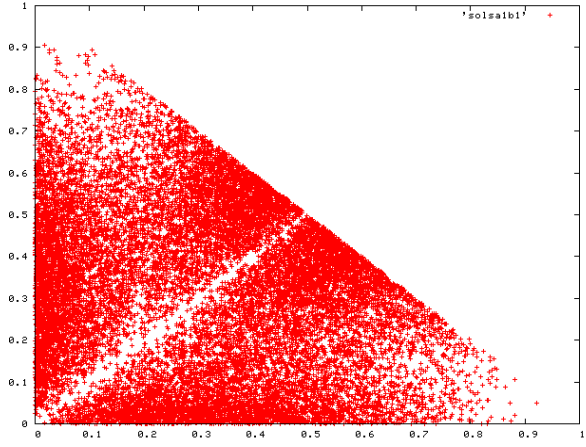}
  \end{center}
  \caption{Decision changes, projected on the plane $a_1, b_1$.}
  \label{solsa1b1}
\end{figure}

The conjunctive rule yields:
$$m_c(\emptyset)=2xy,$$
$$m_c(A)=x+y-2xy-xz = x-m_c(\emptyset)+y(1-x),$$
$$m_c(B)=x+y-xy-2xz = x-m_c(\emptyset)+z(1-x),$$
$$m_c(\Theta)=1-2x-y-z+2xy+2xz.$$

Therefore, as $1-x\geq 0$:

$$m_c(A)>m_c(B) \iff y>z.$$

The PCR yields:
$$m_{PCR}(\emptyset)=0$$
$$m_{PCR}(A)=x-m_c(\emptyset)+y(1-x)+\frac{x^2z}{x+z}+\frac{xy^2}{x+y},$$
$$m_{PCR}(B)= x-m_c(\emptyset)+z(1-x)+\frac{xz^2}{x+z}+\frac{x^2y}{x+y},$$
$$m_{PCR}(\Theta)=1-2x-y-z+2xy+2xz.$$

So, we have:
\begin{eqnarray*}
  (m_{PCR}(A)+m_c(\emptyset)-x)(x+y)(x+z) & = & y(1-x)(x+z)(x+y) \\
  &&~~~~+ x^2z(x+y) +  y^2x(x+z) \\
  & = & y(x+y)(x+z) + x^3(z-y)
\end{eqnarray*}

$$(m_{PCR}(B)+m_c(\emptyset)-x)(x+y)(x+z) = z(x+y)(x+z) - x^3(z-y),$$

$$m_{PCR}(A)>m_{PCR}(B) \iff (y-z)((x+y)(x+y)-2x^3) > 0.$$
As $0\leq x\leq\frac{1}{2}$, we have $2x^3\leq x^2\leq (x+y)(x+z)$. So
$m_{PCR}(A)>m_{PCR}(B)$ if and only if $y>z$.

That shows that the stability of the decision is reached if $a_1=b_1$ for all $a_2$ and $b_2 \in [0,1]$ or by symmetry if $a_2=b_2$ for all $a_1$ and $b_1 \in [0,1]$.

\subsubsection{Case $a_1=b_2$}

In this situation, expert 1 believes $A$ and the expert 2 believes $B$ with the same weight:

\begin{center}
  \begin{tabular}{|l|c|c|c|c|}
    \hline
    & $\emptyset$ & $A$ & $B$ & $\Theta$ \\
    \hline
    Expert 1 & 0 & $x$ & $y$ & $1-x-y$ \\
    \hline
    Expert 2 & 0 & $z$ & $x$ & $1-x-z$ \\
    \hline
  \end{tabular}
\end{center}

\begin{figure}[htb]
  \begin{center}
      \includegraphics[height=6cm]{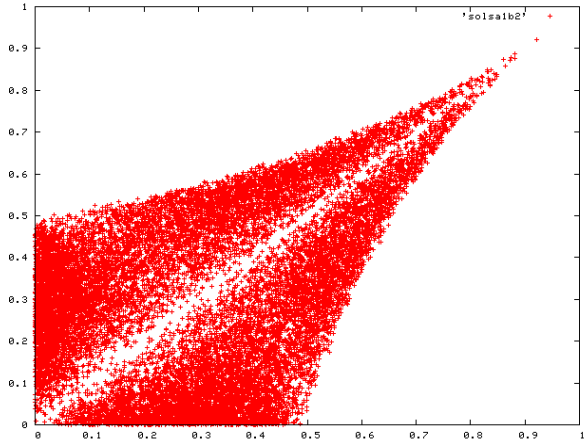}
  \end{center}
  \caption{Decision changes, projected on the plane $a_1, b_2$.}
  \label{solsa1b2}
\end{figure}

The conjunctive rule yields:
$$m_c(\emptyset)=x^2+yz,$$
$$m_c(A)=x+z-xz-m_c(\emptyset)=-x^2+x(1-z)+z(1-y),$$
$$m_c(B)=x+y-xy-m_c(\emptyset)=-x^2+x(1-y)+y(1-z),$$
$$m_c(\Theta)=1+m_c(\emptyset)-2x-y-z+x(y+z).$$
Therefore:
$$m_c(A)>m_c(B) \iff (x-1) (y-z)>0,$$
as $1-x\geq 0$:
$$m_c(A)>m_c(B) \iff y>z.$$

The PCR yields:
$$m_{PCR}(\emptyset)=0,$$
$$m_{PCR}(A)=x+z-xz-m_c(\emptyset)=-x^2+x(1-z)+z(1-y)+\frac{x^3}{2x}+\frac{yz^2}{y+z},$$
$$m_{PCR}(B)=x+y-xy-m_c(\emptyset)=-x^2+x(1-y)+y(1-z)+\frac{x^3}{2x}+\frac{y^2z}{y+z},$$
$$m_{PCR}(\Theta)=1+m_c(\emptyset)-2x-y-z+x(y+z).$$

Therefore:
$$m_{PCR}(A)>m_{PCR}(B) \iff (y-z) \left((x-1) (y+z)-yz\right)>0,$$
as $(x-1)\leq 0$, $(x-1) (y+z)-yz \leq 0$ and:
$$m_{PCR}(A)>m_{PCR}(B) \iff y>z.$$
That shows that the stability of the decision is reached if $a_1=b_2$ for all $a_2$ and $b_1 \in [0,1]$ or by symmetry if $a_2=b_1$ for all $a_1$ and $b_2 \in [0,1]$.

\subsubsection{Case $a_2=1-a_1$}

We can notice that if $a_1+a_2>1$, no change occurs. In this
situation, we have $b_1+b_2<1$, but calculus is still to be done.

\begin{figure}[htb]
  \begin{center}
      \includegraphics[height=6cm]{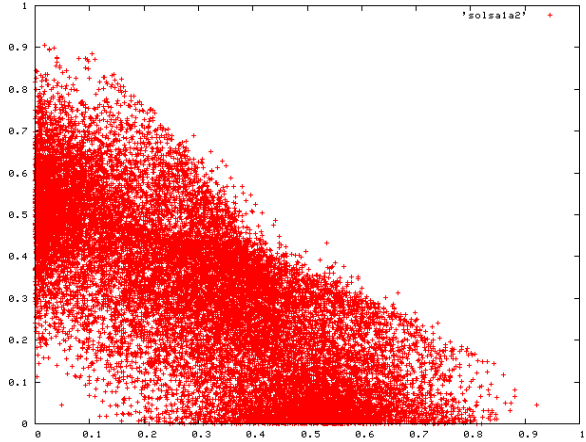}
  \end{center}
  \caption{Decision changes, projected on the plane $a_1, a_2$.}
  \label{solsa1a2}
\end{figure}

In this situation, if $a_2=1-a_1$:

\begin{center}
  \begin{tabular}{|l|c|c|c|c|}
    \hline
    & $\emptyset$ & $A$ & $B$ & $\Theta$ \\
    \hline
    Expert 1 & 0 & $x$ & $y$ & $1-x-y$ \\
    \hline
    Expert 2 & 0 & $1-x$ & $z$ & $x-z$ \\
    \hline
  \end{tabular}
\end{center}

The conjunctive rule yields:
$$m_c(\emptyset)=xz+(1-x)y,$$
$$m_c(A)=1+x^2-x-y+xy-xz,$$
$$m_c(B)=z-yz+xy-xz,$$
$$m_c(\Theta)=-x^2+x+xz-xy+yz-z.$$
Therefore:
$$m_c(A)>m_c(B) \iff 1+x^2-x>y+z-yz,$$
$$\iff x(1-x)>(1-y)(1-z),$$
as $z<x$ and $x<1-y$, $m_c(A)>m_c(B)$ is always true. 

The PCR yields:
$$m_{PCR}(\emptyset)=0,$$
$$m_{PCR}(A)=m_c(A)+\frac{x^2z}{x+z}+\frac{(1-x)^2y}{1-x+y},$$
$$m_{PCR}(B)=m_c(B)+\frac{xz^2}{x+z}+\frac{(1-x)y^2}{1-x+y},$$
$$m_{PCR}(\Theta)=m_c(\Theta).$$

Therefore:
$$m_{PCR}(A)>m_{PCR}(B)$$ is always true.

Indeed we have $m_c(A)>m_c(B)$ is always true and:  
$$\frac{x^2z}{x+z}>\frac{xz^2}{x+z}$$
because $x>z$ and:
$$\frac{(1-x)^2y}{1-x+y}>\frac{(1-x)y^2}{1-x+y}$$
because $1-x>y$.

That shows that the stability of the decision is reached if $a_2=1-a_1$ for all $a_2$ and $a_1 \in [0,1]$ or by symmetry if $a_1=1-a_2$ for all $a_1$ and $a_2 \in [0,1]$.

\section{Conclusion}

In this chapter, we have proposed a study of the combination rules compared in term of decision. A new generalized proportional conflict redistribution (PCR6) rule have been proposed and discussed. We have presented the pro and con of this rule. The PCR6 rule is more intuitive than the PCR5. We have shown on randomly generated data, that there is a difference of decision following the choice of the combination rule (for the non-normalized conjunctive rule, the mixed conjunctive and disjunction rule of Dubois and Prade, the PCR5 rule and the PCR6 rule). We have also proven, on a two experts and two classes case, the changes following the values of the basic belief assignments.  This difference can be very small in percentage and we can not say on these data if it is a significantly difference. We have conducted this comparison on real data in the chapter \cite{Martin06b}.

All this discussion comes from a fine proportional conflict distribution initiated by the consideration of the extension of the discernment space in $D^{\Theta}$. The generalized PCR6 rule can be used on $2^\Theta$ or $D^\Theta$.

\bibliographystyle{plain} 
\bibliography{biblio}

\appendix
\section*{Appendix: Algorithms}

An expert $e$ is an association of a list of focal classes and their
masses. We write $size(e)$ the number of its focal classes. The focal
classes are $e[1]$, $e[2]$, \ldots, $e[size(e)]$. The mass associated
to a class $c$ is $e(c)$, written with paranthesis.

\begin{algorithm}
  \KwData{$n$ experts $ex$: $ex[1] \ldots ex[n]$}
  \KwResult{Fusion of $ex$ by Dubois-Prade method : $edp$}
  \For{$i$ = 1 to $n$}{
    \ForEach{$c$ in $ex[i]$}{
      Append $c$ to $cl[i]$\;
    }
  }
  \ForEach{$ind$ in [1, size($cl[1]$)] $\times$ [1, size($cl[2]$)] $\times$ \ldots $\times$ [1, size($cl[n]$)]}{
    $s$ $\gets$ $\Theta$\;
    \For{$i$ = 1 to $n$}{
      $s$ $\gets$ $s\cap cl[i][ind[i]]$\;
    }
    \If{$s$ = $\emptyset$}{
      $lconf$ $\gets$ 1\;
      $u$ $\gets$ $\emptyset$\;
      \For{$i$ = 1 to $n$}{
        $u$ $\gets$ $p\cup cl[i][ind[i]]$\;
        $lconf$ $\gets$ $lconf \times ex[i](cl[i][ind[i]])$\;
      }
      $edp(u) \gets edp(u)+lconf$\;
    }
    \Else{
      $lconf$ $\gets$ 1\;
      \For{$i$ = 1 to $n$}{
        $lconf$ $\gets$ $lconf \times ex[i](cl[i][ind[i]])$\;
      }
      $edp(s) \gets edp(s)+lconf$\;
    }
  }
\end{algorithm}

\begin{algorithm}
  \KwData{$n$ experts $ex$: $ex[1] \ldots ex[n]$}
  \KwResult{Fusion of $ex$ by PCR5 method : $ep$}
  \For{$i$ = 1 to $n$}{
    \ForEach{$c$ in $ex[i]$}{
      Append $c$ to $cl[i]$\;
    }
  }
  \ForEach{$ind$ in [1, size($cl[1]$)] $\times$ [1, size($cl[2]$)] $\times$ \ldots $\times$ [1, size($cl[n]$)]}{
    $s$ $\gets$ $\Theta$\;
    \For{$i$ = 1 to $n$}{
      $s$ $\gets$ $s\cap cl[i][ind[i]]$\;
    }
    \If{$s$ = $\emptyset$}{
      $lconf$ $\gets$ 1\;
      $el$ is an empty expert\;
      \For{$i$ = 1 to $n$}{
        $lconf$ $\gets$ $lconf \times ex[i](cl[i][ind[i]])$\;
        \If{$cl[i][ind[i]]$ in $el$}{
          $el(cl[i][ind[i]]) \gets el(cl[i][ind[i]]) * ex[i](cl[i][ind[i]])$\;
        }
        \Else{
          $el(cl[i][ind[i]]) \gets ex[i](cl[i][ind[i]])$\;
        }
      }
      \For{$c$ in $el$}{
        $sum \gets sum + el(c)$\;
      }
      \For{$c$ in $el$}{
        $ep(c) \gets ep(c) + g(el(c))*lconf/sum$\;
      }
    }
    \Else{
      $lconf$ $\gets$ 1\;
      \For{$i$ = 1 to $n$}{
        $lconf$ $\gets$ $lconf \times ex[i](cl[i][ind[i]])$\;
      }
      $ep(s) \gets ep(s)+lconf$\;
    }
  }
\end{algorithm}

\begin{algorithm}
  \KwData{$n$ experts $ex$: $ex[1] \ldots ex[n]$}
  \KwResult{Fusion of $ex$ by PCR6 method : $ep$}
  \For{$i$ = 1 to $n$}{
    \ForEach{$c$ in $ex[i]$}{
      Append $c$ to $cl[i]$\;
    }
  }
  \ForEach{$ind$ in [1, size($cl[1]$)] $\times$ [1, size($cl[2]$)] $\times$ \ldots $\times$ [1, size($cl[n]$)]}{
    $s$ $\gets$ $\Theta$\;
    \For{$i$ = 1 to $n$}{
      $s$ $\gets$ $s\cap cl[i][ind[i]]$\;
    }
    \If{$s$ = $\emptyset$}{
      $lconf$ $\gets$ 1\;
      $sum$ $\gets$ 0\;
      \For{$i$ = 1 to $n$}{
        $lconf$ $\gets$ $lconf \times ex[i](cl[i][ind[i]])$\;
        $sum$ $\gets$ $sum + ex[i](cl[i][ind[i]])$\;        
      }
      \For{$i$ = 1 to $n$}{
        $ep(ex[i][ind[i]]) \gets ep(ex[i][ind[i]]) + ex[i](cl[i][ind[i]])*lconf/sum$\;
      }
    }
    \Else{
      $lconf$ $\gets$ 1\;
      \For{$i$ = 1 to $n$}{
        $lconf$ $\gets$ $lconf \times ex[i](cl[i][ind[i]])$\;
      }
      $ep(s) \gets ep(s)+lconf$\;
    }
  }
\end{algorithm}

\begin{algorithm}
  \KwData{$n$ experts $ex$: $ex[1] \ldots ex[n]$}
  \KwData{A non-decreasing positive function $f$}
  \KwResult{Fusion of $ex$ by PCR6$_f$ method : $ep$}
  \For{$i$ = 1 to $n$}{
    \ForEach{$c$ in $ex[i]$}{
      Append $c$ to $cl[i]$\;
    }
  }
  \ForEach{$ind$ in [1, size($cl[1]$)] $\times$ [1, size($cl[2]$)] $\times$ \ldots $\times$ [1, size($cl[n]$)]}{
    $s$ $\gets$ $\Theta$\;
    \For{$i$ = 1 to $n$}{
      $s$ $\gets$ $s\cap cl[i][ind[i]]$\;
    }
    \If{$s$ = $\emptyset$}{
      $lconf$ $\gets$ 1\;
      $sum$ $\gets$ 0\;
      \For{$i$ = 1 to $n$}{
        $lconf$ $\gets$ $lconf \times ex[i](cl[i][ind[i]])$\;
        $sum$ $\gets$ $sum + f(ex[i](cl[i][ind[i]]))$\;        
      }
      \For{$i$ = 1 to $n$}{
        $ep(ex[i][ind[i]]) \gets ep(ex[i][ind[i]]) + f(ex[i](cl[i][ind[i]]))*lconf/sum$\;
      }
    }
    \Else{
      $lconf$ $\gets$ 1\;
      \For{$i$ = 1 to $n$}{
        $lconf$ $\gets$ $lconf \times ex[i](cl[i][ind[i]])$\;
      }
      $ep(s) \gets ep(s)+lconf$\;
    }
  }
\end{algorithm}

\begin{algorithm}
  \KwData{$n$ experts $ex$: $ex[1] \ldots ex[n]$}
  \KwData{A non-decreasing positive function $g$}
  \KwResult{Fusion of $ex$ by PCR6$_g$ method : $ep$}
  \For{$i$ = 1 to $n$}{
    \ForEach{$c$ in $ex[i]$}{
      Append $c$ to $cl[i]$\;
    }
  }
  \ForEach{$ind$ in [1, size($cl[1]$)] $\times$ [1, size($cl[2]$)] $\times$ \ldots $\times$ [1, size($cl[n]$)]}{
    $s$ $\gets$ $\Theta$\;
    \For{$i$ = 1 to $n$}{
      $s$ $\gets$ $s\cap cl[i][ind[i]]$\;
    }
    \If{$s$ = $\emptyset$}{
      $lconf$ $\gets$ 1\;
      $el$ is an empty expert\;
      \For{$i$ = 1 to $n$}{
        $lconf$ $\gets$ $lconf \times ex[i](cl[i][ind[i]])$\;
        $el(cl[i][ind[i]]) \gets el(cl[i][ind[i]]) + ex[i](cl[i][ind[i]])$\;
      }
      $sum$ $\gets$ 0\;
      \For{$c$ in $el$}{
        $sum \gets sum + g(el(c))$\;
      }
      \For{$c$ in $el$}{
        $ep(c) \gets ep(c) + g(el(c))*lconf/sum$\;
      }
    }
    \Else{
      $lconf$ $\gets$ 1\;
      \For{$i$ = 1 to $n$}{
        $lconf$ $\gets$ $lconf \times ex[i](cl[i][ind[i]])$\;
      }
      $ep(s) \gets ep(s)+lconf$\;
    }
  }
\end{algorithm}

\end{document}